\documentclass[letterpaper, 10 pt, conference]{ieeeconf}  

\IEEEoverridecommandlockouts                              

\usepackage{graphicx} 

\usepackage{preamble}
\usepackage{multirow}
\usepackage{booktabs}
\usepackage{algorithm}
\usepackage[noend]{algpseudocode}
\usepackage{bbm}
\usepackage[table]{xcolor}
\newcommand{\best}[1]{\cellcolor{green!20}{#1}}
\newcommand{\secbest}[1]{\cellcolor{yellow!25}{#1}}

\theoremstyle{remark}     
\newtheorem{problem}{Problem}

\algrenewcommand\algorithmicensure{\textbf{Output:}}
\algrenewcommand\algorithmiccomment[1]{\hfill\textcolor{green!50!black}{$\triangleright$~#1}}

\DeclareMathOperator*{\KL}{KL}
\definecolor{darkgreen}{rgb}{0.0, 0.5, 0.0}

\title{\LARGE \bf
Distribution Estimation for Global Data Association via Approximate Bayesian Inference
}

\author{Yixuan Jia\textsuperscript{1}, Mason B. Peterson\textsuperscript{1}, Qingyuan Li\textsuperscript{1}, Yulun Tian\textsuperscript{2}, Jonathan P. How\textsuperscript{1}
\thanks{This work is supported by ARL DCIST under Cooperative Agreement Number W911NF-17-2-0181.}
\thanks{\textsuperscript{1}Massachusetts Institute of Technology, Cambridge, MA 02139, USA. \{\texttt{yixuany, masonbp, andyli27, jhow}\}\texttt{@mit.edu}.}%
\thanks{\textsuperscript{2}Robotics Department, University of Michigan, MI 48109, USA. \texttt{yulunt@umich.edu}.}%
}

\usepackage[noadjust,sort,compress]{cite}

\makeatletter
\let\NAT@parse\undefined
\makeatother

\usepackage[colorlinks=true,linkcolor=black,citecolor=black,urlcolor=blue]{hyperref}
\usepackage[nameinlink]{cleveref}
\crefname{equation}{Eq.}{Eqs.}
\Crefname{equation}{Eq.}{Eqs.}

\crefname{problem}{Problem}{Problems}
\Crefname{problem}{Problem}{Problems}
\crefname{figure}{Fig.}{Figs.}
\Crefname{figure}{Fig.}{Figs.}

\makeatletter
\let\orglabel\label
\renewcommand{\label}[1]{\orglabel{#1}\hypertarget{#1}{}}
\makeatother

\begin{document}

\maketitle
\thispagestyle{empty}
\pagestyle{empty}

\begin{abstract}

Global data association is an essential prerequisite for robot operation in environments seen at different times or by different robots.
Repetitive or symmetric data creates significant challenges for existing methods, which typically rely on maximum likelihood estimation or maximum consensus to produce a {\em single} set of associations. However, in ambiguous scenarios, the 
distribution of solutions to global data association problems is often highly multimodal,
and such single-solution approaches frequently fail.
In this work, we introduce a data association framework that leverages approximate Bayesian inference to capture multiple solution modes to the data association problem, thereby avoiding premature commitment to a single solution under ambiguity. 
Our approach represents hypothetical solutions as particles that evolve according to a deterministic or randomized update rule to cover the modes of the underlying solution distribution.
Furthermore, we show that our method can incorporate optimization constraints imposed by the data association formulation and directly benefit from GPU-parallelized optimization.
Extensive simulated and real-world experiments with highly ambiguous data show that our method correctly estimates the distribution over transformations when registering point clouds or object maps.

\end{abstract}

\section{INTRODUCTION}
Data association is essential in many robotic applications, enabling key perception technologies such as dynamic object tracking \cite{strecke2019fusion,peterson2025tcaff,reid2003algorithm} and simultaneous localization and mapping (SLAM) \cite{bowman2017probabilistic,doherty2020probabilistic,tian2022kimera}. In these scenarios, robots must recognize when an object or feature they are currently observing is the same as something they (or another robot) may have seen from a different perspective. 
Without correct data association, the environment representation may be inconsistent, leading to undesirable behaviors in downstream tasks (e.g., incorrect associations in loop closure detection can lead to dramatically distorted maps~\cite{tian2022kimera}).

\begin{figure}[t]
    \centering
    \includegraphics[width=0.48\textwidth]{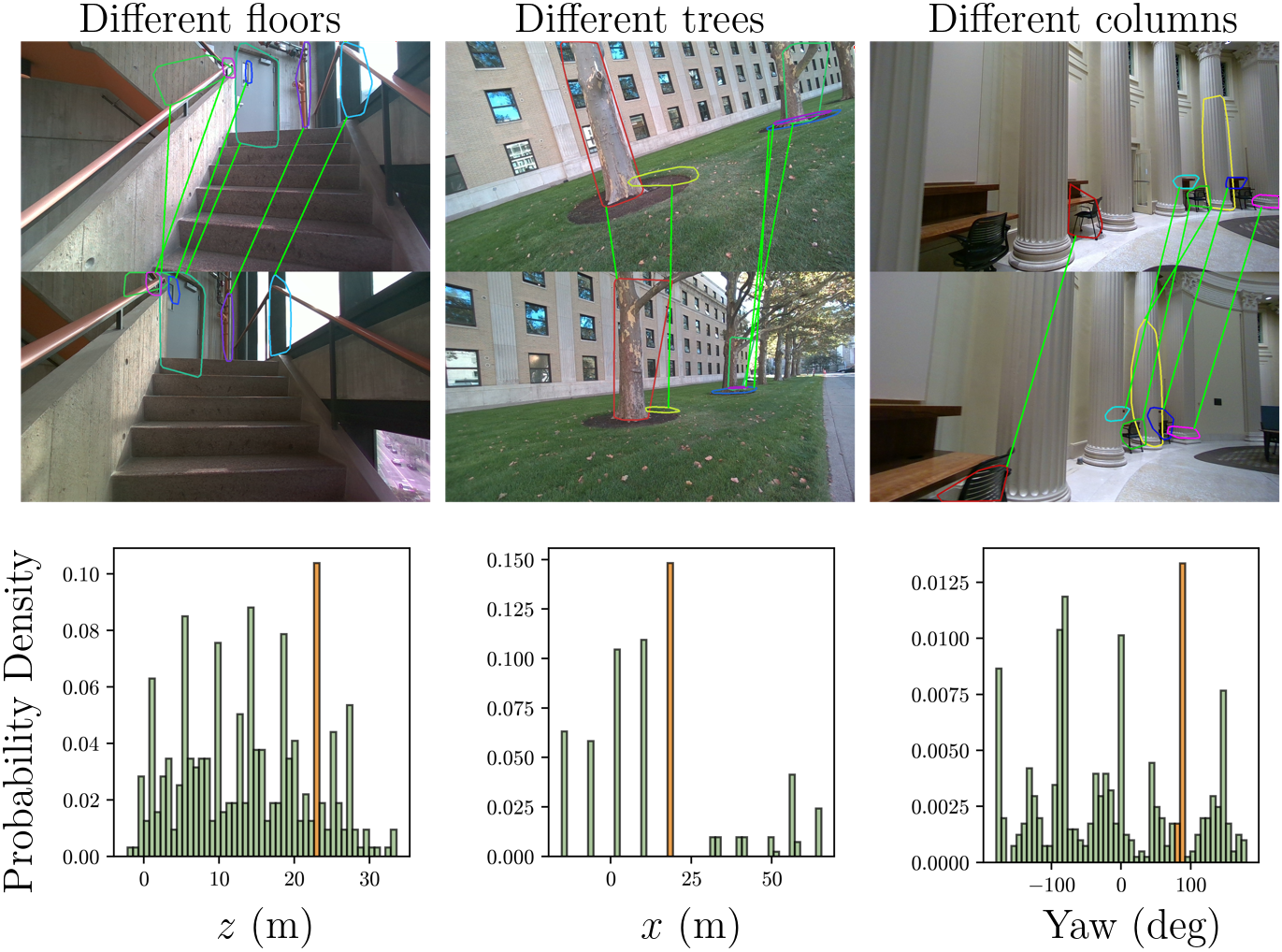}
    \caption{Symmetric and repetitive structures are common in human environments and induce perceptual aliasing. To localize, robots must explicitly model the uncertainty from ambiguities inherent in those environments. 
    The top row shows several such scenarios, where the visualized object associations are perceptually similar but incorrect, necessitating a multimodal representation.
    The bottom row shows the multimodal distribution generated by our proposed method.
    The mode visualized in the top row is colored orange.
    }
    \label{fig:demo}
\end{figure}

This work considers \emph{global} data association, which aims to find 
pairwise correspondences for registering 
two point clouds \cite{kim2018scan}, object-level maps \cite{peterson2024roman}, or scene graphs \cite{chang2023hydra} \emph{without an initial guess}.
While significant progress has been made towards point-to-point data association \cite{fischler1981random,yang2020teaser,mangelson2018pairwise}, most existing methods formulate data association using maximum likelihood estimation 
or maximum consensus, returning a {\it single} solution representing the most likely relative pose estimate or most consistent set of data associations.
However, both approaches can fail in the presence of ambiguous data, where multiple likely solutions coexist and cannot be distinguished without additional observations. Such ambiguities are common in human environments, which often contain symmetrical or repetitive structures. 
In these scenarios, the success of the solver may be determined purely by the optimization initialization or the realization of measurement noise.
For example, consider a robot trying to localize itself from a reference map within the staircase shown in~\Cref{fig:demo}.
The repetitive nature of each floor of the staircase results in many likely sets of object associations, yielding a multimodal distribution over the robot pose estimate. 
If the robot were to choose the most likely option, it would believe that it was two floors above its actual location.


\raggedbottom


Currently, there are mainly three applicable approaches to avoid incorrect estimations in ambiguous scenarios: (1) outlier rejection, (2) probabilistic data association, and (3) probabilistic pose estimation. 
Existing outlier rejection methods \cite{sunderhauf2012switchable,lee2013robust,mangelson2018pairwise,yang2020graduated} typically assume that all inlier measurements are mutually consistent while all outliers are uncorrelated. As a result, they do not work well when measurements form multiple consistent clusters and the resulting solution distribution has multiple modes. 
On the other hand, probabilistic data association methods \cite{mu2016slam,bowman2017probabilistic,doherty2020probabilistic,hsiao2019mh} model the distributions of associations in a factor graph 
by introducing new measurement models and factors. 
However, 
these methods
primarily address local ambiguities rather than global ambiguities that are inherent in highly symmetric or repetitive environments.
Another line of work augments existing methods such as ICP with probabilistic methods \cite{maken2020estimating,maken2021stein} to estimate uncertainty in pose estimation. 
However, as shown in our experiments, such approaches struggle to estimate highly non-uniform distributions.
\textbf{Contributions.} In this work, we address the limitations of the aforementioned work by proposing novel algorithms for \emph{multimodal} global data association.
Our core insight is to view data association from the lens of approximate Bayesian inference, by interpreting the former's objective function as log likelihood defined over sets of candidate associations.
Building on this insight, we develop deterministic and randomized multimodal global data association algorithms that connect with two well-established Bayesian inference frameworks.
Specifically, our approach models hypothetical solutions as a set of particles that are updated using approximate-inference-based update rules. 
We show that the proposed method can handle difficulties specific to the optimization formulation of global data association and directly benefit from modern GPU-parallelized optimization solvers (code will be released).
Experimentally, we perform extensive comparison studies against existing approaches and show the proposed approach can capture both highly peaked and highly uniform solution distributions. 

\section{Related Works}
{\bf Data Association}
In robotics, data association refers to the task of matching observations of the environment across agents or time. 
Specifically, we address methods that consider distributions over solutions to the data association problem, rather than methods that commit to a single hypothesis (i.e., a single most likely pose estimate or a single set of data associations). 

Although recent work has begun to address multimodal data association problems,
most of these efforts concentrate on SLAM backend design, particularly in the context of factor graph optimization. 
For example, work on probabilistic data association \cite{mu2016slam,bowman2017probabilistic,doherty2020probabilistic} focuses on representing distributions of associations in the factor graph by introducing new measurement models. 
However, this line of work often assumes good initial guesses provided by the frontend and usually focuses on extracting the most likely solution instead of estimating the full distribution.
For example, Mu et al. \cite{mu2016slam} introduce a nonparametric factor graph formulation and use a Dirichlet process to represent the prior over data associations. 
However, a small variance assumption is made on the posterior distribution of data associations and the posterior distribution is assumed to have a unique maximal value. 
Bowman et al. \cite{bowman2017probabilistic} instead maintain probabilistic weights of potential data associations via expectation maximization (EM). However, the hypotheses are blended through expectation, thus not representing the full distribution. 
Instead of averaging via EM, Doherty et al. \cite{doherty2020probabilistic} use a max-mixture model with max-marginalization to allow for switching between hypotheses.
%
Another approach uses Multi-Hypothesis Tracking (MHT) \cite{reid2003algorithm,hsiao2019mh,peterson2025tcaff} to track multiple possible solutions in data association problems. For example, Hsiao and Kaess \cite{hsiao2019mh} propose an extension to classic factor graph SLAM that tracks multiple hypotheses at the back-end, assuming the front-end provides the hypotheses. 
However, MHT is typically computationally expensive due to the exponential growth of the hypothesis tree. 

The most relevant line of work to ours is \cite{maken2020estimating,maken2021stein}. In \cite{maken2020estimating}, Maken et al. develop a variant of Iterative Closest Point (ICP), termed Bayesian ICP, which uses stochastic gradient Langevin dynamics \cite{welling2011bayesian} to obtain samples from the posterior distributions of relative poses between point clouds. 
Because Bayesian ICP is computationally expensive and not parallelized, the authors then propose Stein ICP \cite{maken2021stein}, which employs Stein variational gradient descent (SVGD) to approximate the posterior distribution more efficiently, achieving better empirical performance. 

{\bf Bayesian Inference in Robotics}
Approximating or sampling from a distribution defined by an unnormalized density function is one of the core problems in modern statistics. 
Two major classes of methods are Markov Chain Monte Carlo (MCMC) \cite{robert1999monte,neal1993probabilistic,gilks1995markov} and Variational Inference (VI) \cite{jordan1999introduction,wainwright2008graphical,blei2017variational}. 
While MCMC constructs a Markov chain to draw samples from the posterior distribution, VI approximates the posterior distribution directly by selecting a tractable family of distributions and identifying the member that minimizes the KL divergence to the posterior. 
Among VI algorithms, Stein variational gradient descent (SVGD) has recently attracted increasing attention in robotics, where applications ranging from state estimation \cite{maken2021stein,maken2022stein,koide2024megaparticles} to planning \cite{lambert2020stein,lambert2022,lee2024,power2024constrained,pavlasek2024}. Motivated by these successes, we develop a variant of our method based on SVGD and investigate its performance in real-world global data association problems.
Among MCMC algorithms, Langevin dynamics (referred to as the Unadjusted Langevin Algorithm when discretized) has been extensively studied both in the context of stochastic optimization as a way to escape local minima \cite{geman1986diffusions,gelfand1991recursive,raginsky2017non} and in Bayesian inference where it serves as a method for approximating the posterior distributions \cite{roberts1996exponential,durmus2017nonasymptotic,dalalyan2017theoretical,vempala2019rapid}. 
We develop a second variant of our algorithms based on Langevin dynamics and show that the dual role of Langevin dynamics in both global optimization and posterior approximation makes it surprisingly effective for approximating posterior distributions in global data association problems.

\section{Preliminaries}
Given two ordered lists of points $\mS, \mT$, we can form a list of potential associations. 
For example, if each point in $\mS$ can be associated with each point in $\mT$, we can obtain $|\mS| \times |\mT|$ potential associations. 
The association between point $s \in \mS$ and point $t \in \mT$ is denoted as $(s, t)$.
To ease the burden of notation, we will use the index of a point interchangeably with its coordinate, e.g.,
$s$ will denote both the $s$-th point in $\mS$ as well as the coordinate of the $s$-th point in $\mS$ expressed in a local reference frame (e.g. robot's sensor frame). 
We use $\|\cdot\|_2$ to denote the 2-norm in Euclidean space.

{\bf CLIPPER}
Given $n$ potential associations (e.g., $ n = |\mS| \times |\mT|$), we
can form an affinity matrix $M \in \R^{n \times n}_+$, which is a symmetric matrix with entries in $[0, 1]$. 
Let \(i = (s_i, t_i)\) and \(j = (s_j, t_j)\),  
CLIPPER \cite{lusk2021clipper} then defines 
\(M[i,j] = e^{-d((s_i, t_i),(s_j, t_j))^2 / (2\sigma^2)}\) 
if \(d((s_i, t_i),(s_j, t_j)) < \varepsilon\), and \(M[i,j] = 0\) otherwise,
where $d((s_i, t_i), (s_j, t_j)) = \left| \|s_i - s_j\|_2 - \|t_i - t_j\|_2 \right|$. 
Intuitively, $d(\cdot, \cdot)$ measures the violation of geometric consistency, since a consistent set of associations should preserve the relative distances of points. Then the problem of finding the maximum consistent set of associations can be formulated as follows:
%
\begin{problem}\label{prob:clipper_orig}
\begin{equation*}
    \begin{split}
        \underset{u \in \{0, 1\}^n}{\max}  
        &\frac{{u}^\top {M} {u}}{{u}^\top {u}} \\
        \mathrm{subject} ~ \mathrm{to} \quad & u_i u_j = 0 \;\; \text{if}\;\; M[i,j] = 0, \; \forall i,j.
    \end{split}
\end{equation*}
\end{problem}
\noindent
The matrix $M$ can be viewed 
as the sum of a weighted adjacency matrix and the identity matrix, 
where the weighted adjacency matrix defines a graph called
the {\it consistency graph}, 
wherein nodes represent potential associations and edges connect nodes only if they are consistent.
In the case where $M$ is binary, \Cref{prob:clipper_orig} is equivalent to finding the maximum clique in the consistency graph, which is expensive to obtain for large graphs. Therefore, CLIPPER solves a relaxed problem, described as follows.
Define another matrix \(M_d\) such that 
\(M_d[i, j] = M[i, j]\) if \(M[i, j] \neq 0\), and \(M_d[i, j] = -d\) otherwise, where $d \in \R_+$. 
Intuitively, if two associations are deemed inconsistent, the corresponding entry in $M_d$ will incur penalties if selected.
Then, inspired by \cite{belachew2017solving}, CLIPPER solves an optimization problem of the form:
\begin{problem}\label{prob:clipper}
\begin{equation*}
\begin{aligned}
    &\max_{u \in \R^n_+} \quad u^TM_d u \\
    &\mathrm{subject} ~ \mathrm{to} \quad \|u\|_2 \le 1.
\end{aligned}
\end{equation*}
\end{problem} 
\noindent 
The optima lie on the boundary of $\|u\|_2 \le 1$ \cite{lusk2021clipper}, so we can equivalently write $u \in \R^n_+ \cap \S^{n-1} $ and remove the inequality constraint. Moreover, as remarked in \cite{lusk2021clipper}, when $d \ge n$, the (local) optima of \Cref{prob:clipper} correspond to the (local) optima of the original \Cref{prob:clipper_orig}. The local optima correspond to maximal cliques in the consistency graph while the global optimum corresponds to the maximum clique.

{\bf Stein Variational Gradient Descent}
SVGD aims to approximate a target distribution with an unnormalized density function $p: \R^n \to \R$ using a set of particles $X = \{x_i\}_{i=1}^N, ~x_i \in \R^n ~\forall i$. Denoting the distribution represented by the set of particles $X$ as $q_X$, the goal is to find $X$ such that $\KL(q_X\parallel p)$ is minimized. Suppose we would like to iteratively update the particles via a map $T(x)$ that takes the form $x + \alpha \phi(x)$ where $\alpha > 0$ is the step size.
The optimal perturbation direction \(\phi(\cdot)\) is 
\(\max_{\|\phi\|_{\mathcal{H}} \leq 1} \big\{-\tfrac{d}{d\alpha} \KL(Tq_X \parallel p)\big|_{\alpha=0}\big\}\), 
where $Tq_X$ denotes the pushforward of $q_X$ under the map $T$ and  \(\mathcal{H}\) is a reproducing kernel Hilbert space \cite{liu2017stein}.
 Intuitively, $\phi(\cdot)$ is the functional gradient of the $\KL$ divergence with respect to the particles. A key result from \cite{liu2016stein} shows that $\phi(\cdot)$ takes a closed-form expression:
\begin{equation}\label{eq:svgd}
    \phi(x_i) = \frac{1}{N} \sum_{j=1}^N \nabla \log p(x_j)\, k(x_i, x_j) 
    + \nabla_{x_j} k(x_i, x_j),
\end{equation}
where $k(\cdot, \cdot)$ is a kernel specified by the user.
Intuitively, SVGD works by pushing the particles to high-density regions via the $\nabla \log p(\cdot)$ term, while pulling them apart from each other to encourage discovering different modes via the repulsive term $\nabla k(\cdot, \cdot)$. Note that the repulsive term is deterministic, which makes SVGD a deterministic algorithm. With only a single particle, SVGD reduces to exact  maximum a posteriori (MAP) optimization \cite{liu2017stein}.

{\bf Langevin Dynamics}
Given an unnormalized density function $p: \R^n \to \R$, Langevin dynamics (LD) can be applied to draw samples from the distribution defined by $p(\cdot)$ by iteratively updating the particle \cite{liu2017stein}:
\begin{equation}
   x_{t+1} \gets x_t + \alpha \nabla \log p(x_t) + 2 \sqrt{\alpha} \xi_t, ~ \xi_t \sim \mN(0, 1). 
   \label{eq:langevin}
\end{equation}
Intuitively, the second term on the right side encourages particles to cover regions with higher density, and the third term encourages particles to spread out so that more modes can be discovered. Unlike SVGD, LD is a stochastic algorithm due to the random noise term. Moreover, the computation complexity is reduced by removing the $O(n^2)$ computation needed to evaluate the kernel terms in SVGD.
This update can be viewed as a discrete version of an It\^o diffusion $dx_t = -\nabla \log p(x_t) dt + 2 dB_t$ where $B_t$ is a standard Brownian motion. It is well-known that the KL divergence between the density of $x_t$ and $p(\cdot)$ decays at a rate equal to the Fisher divergence \cite{liu2017stein}.



\section{Proposed Method}
In this section, we introduce our proposed methods. The general strategy is to view \Cref{prob:clipper} from a probabilistic perspective in which the objective becomes drawing samples from a target distribution. Then, the approximate Bayesian inference techniques introduced in the previous section can be applied to approximate the target distribution.
Referring to notations in \Cref{prob:clipper}, let $F_d(u) = u^T M_d u$. The data association problem can be viewed as obtaining samples from the posterior distribution (assuming a uniform prior): 
\begin{equation}\label{eq:langevin_target_density}
    \begin{aligned}
        p(u | \mS, \mT)  \propto p(\mS, \mT | u) \propto \exp(F_d(u)), ~ u \in \R^n_+ \cap \S^{n-1}.
    \end{aligned}
\end{equation}
This type of reformulation of an optimization problem into an inference problem is common in trajectory optimization and control literature \cite{toussaint2009robot,
okada2020variational}
. 
Since $p(u | \mS, \mT)$ rarely admits a closed-form expression, our goal is to approximate $p(u | \mS, \mT)$ with a distribution $q(u)$ that is tractable for computation. 




In this work, we propose two variants of our method: {\bf Stein CLIPPER} based on SVGD and  {\bf Langevin CLIPPER} based on Langevin dynamics.
Both algorithms update particles in parallel, enabling efficient implementation on GPUs.

{\bf Stein CLIPPER} Recall the update direction in SVGD given by \Cref{eq:svgd}.
To compute the update, we need the gradient of the log posterior (i.e. the score) as well as a kernel function. 
Since $p(u | \mS, \mT) \propto \exp(F_d(u))$, $\log p(u | \mS, \mT) \propto F_d(u)$. 
Because the score is independent of the normalization constant, we have:
\begin{equation}\label{eq:grad_log_prob}
    \nabla_u \log p(u | \mS, \mT) = \nabla_u F_d(u) = 2 u^TM_d.
\end{equation}
Note that we adopt the convention that the gradient of a function $f : \R^n \to \R^m$ is an $m \times n$ matrix \cite{munkres2018analysis}.
We choose the widely used radial basis function (RBF) kernel \cite{aronszajn1950theory}, which takes the form 
$
k(x, x') = e^{-\|x - x'\|_2^2/(2\sigma_k^2)}.
$

See \Cref{alg:stein_clipper} for the overall Stein CLIPPER approach. Line 1 initializes $n_p$ particles from $\mathrm{Uniform}([0, 1]^n)$ that each represents a possible set of associations (i.e. $u$ in 
\Cref{prob:clipper}). We use a homotopy method  \cite{lusk2021clipper} to gradually increase $d$ (see \Cref{prob:clipper_orig}). We initialize $\Delta d$ to be the maximum eigenvalue of $M$ and increase $d$ by $\Delta d$ for every outer iteration, until $d \ge n$, at which point the optima of \Cref{prob:clipper_orig} should be recovered (see \Cref{prob:clipper} and the comments immediately after it). Empirically, we find setting $\Delta d = \lambda_1(M)$ offers good trade-offs between speed and convergence without using a problem-specific parameter. This homotopy step is important for both the original CLIPPER (to avoid being trapped in local minima) and Stein CLIPPER (to prevent particle degeneracy). 
The inner loop (Lines 6 - 10) consists of the main optimization steps. Intuitively, for each fixed $d$, we solve \Cref{prob:clipper} via gradient ascent where the gradient is computed via \Cref{eq:svgd} and adjusted by the adaptive gradient algorithm AdaGrad \cite{duchi2011adaptive}. AdaGrad adjusts and rescales the learning rate for each particle automatically. Empirically, we find AdaGrad very important for returning diverse particles and preventing particle degeneracy. After the update, we perform a projection step to project each particle back to $\R^n_+ \cap \S^{n-1}$.
Finally, a clique is extracted from each particle by taking the associations corresponding to their top $\hat{\omega} = \mathrm{round}(u^\top M_d u)$ entries as done in \cite{lusk2021clipper}. 

\begin{algorithm}[t]
\caption{Stein CLIPPER}
\label{alg:stein_clipper}
\begin{algorithmic}[1]
\Require Affinity matrix $M \in [0,1]^{n \times n}$, step size $\alpha$, kernel $k(\cdot, ~\cdot)$
\State $\theta \gets \mathrm{rand}(n_p, n)$  \Comment{initialize uniformly in $[0, 1]$}
\State $d \gets 0$ 
\State $\Delta d \gets \lambda_1(M)$ \Comment{$\lambda_1(M)$: maximum eigenvalue of $M$}
\While{$d < n$}
    \State $d \gets d + \Delta d$
    \While{max iterations not reached}
        \State $M_d \gets M - d C$
        \State {Compute $\phi(\theta_i)$ via \Cref{eq:svgd} for each $i \in [n_p]$}
        \State {$\theta \gets \theta + \alpha \cdot \mathrm{AdaGrad}(\phi(\theta))$}
        \State $\theta \gets \max(\theta / \|\theta\|,\ 0)$ 
    \EndWhile
\EndWhile
\State Extract a clique from each $\theta_i$ by taking the top $\hat{\omega}_i = \mathrm{round}({\theta_i}^\top M_d \theta_i)$ entries of $\theta_i$
\end{algorithmic}
\end{algorithm}


{\setlength{\textfloatsep}{0pt}%
 \setlength{\floatsep}{0pt}%
 \setlength{\intextsep}{0pt}%
\begin{algorithm}[t]
\caption{Langevin CLIPPER}
\label{alg:langevin_clipper}
\begin{algorithmic}[1]
\Require Affinity matrix $M \in [0,1]^{n \times n}$, step size $\alpha$
\State $\theta \gets \mathrm{rand}(n_p, n)$ \Comment{initialize uniformly in $[0, 1]$}
\State $M_d \gets M - n C$
\While{max iterations not reached}
    \State {Compute $\phi(\theta_i)$ via \Cref{eq:langevin} for each $i \in [n_p]$}
    \State $\theta \gets \theta + \alpha \cdot \mathrm{AdaGrad}(\phi(\theta))$ 
    \State $\theta \gets \max(\theta / \|\theta\|,\ 0)$  
\EndWhile
\State Extract a clique from each $\theta_i$ by taking the top $\hat{\omega}_i = \mathrm{round}({\theta_i}^\top M_d \theta_i)$ entries of $\theta_i$
\end{algorithmic}
\end{algorithm}
}

{\bf Langevin CLIPPER}
Recall that the update step from Langevin dynamics is given by \Cref{eq:langevin}. Thus we only need the expression for $ \nabla_u \log p(u | \mS, \mT)$ to perform the update, which was derived in \Cref{eq:grad_log_prob}. 

The overall Langevin CLIPPER algorithm is recorded in \Cref{alg:langevin_clipper}. Note that $d$ is directly set to $n$ (Line 2) instead of gradually being increased via the homotopy method. 
We find that Langevin CLIPPER is able to reliably capture the full distribution without the homotopy step, avoiding the particle degeneracy observed with SVGD. 
We hypothesize that this is due to a rigorously motivated relative noise scale with respect to the score ($\sqrt{\alpha}$ vs. ${\alpha}$) \cite{roberts1996exponential}, which automatically adjusts the noise scale to prevent overshooting while still being able to spread out the particles enough to escape the basin of local minima. 
In contrast, the kernel term in SVGD has to be manually selected and properly tuned.
Lines 3 - 7 iteratively update the particles by moving them in the direction given by \Cref{eq:langevin}. Again, AdaGrad helps rescale the step size for each particle and a clique is extracted by taking the top $\hat{\omega}_i$ entries of $\theta_i$ for each $i \in \{1, \dots, n_p\}$.


\begin{remark} \label{rem:bk}
Theoretically, when the affinity matrix $M$ is binary, the modes of the distribution obtained from the proposed methods correspond to the local optima of \Cref{prob:clipper}, which correspond to maximal cliques in the consistency graph. 
Enumerating all maximal cliques using classical approaches such as the Bron-Kerbosch algorithm has a worst-case run time $O(3^{n/3})$ \cite{bron1973algorithm}, which is intractable for most data association applications (e.g., associating $|\mathcal{S}| = 40$ points to $|\mathcal{T}| = 40$ points would result in $n = 40^2 = 1600$). 
The proposed methods enable the approximate enumeration of maximal cliques in a much more tractable manner, as will be shown in \Cref{subsec:bk_result}.
\end{remark}

\section{Simulation Experiments}

Our simulation experiments consist of two types of setups: 
\begin{itemize}
    \item {\bf Simulated Object Map Registration} presented in~\Cref{sec:sim-obj-map-reg}, where the goal is to evaluate the proposed methods under perfect sensing (no noise or outlier objects). These maps are motivated by real-world examples, as shown in \Cref{fig:benchmark:exp_example}. 
    \item {\bf Point Cloud Registration} presented in ~\Cref{sec:sim_obj_pcd}, where we show that the proposed methods can also be applied to point cloud registration tasks as studied in \cite{maken2021stein}.
\end{itemize}
All experiments are performed on a desktop with an RTX 3090 GPU and an i9 CPU with 20 cores.
\subsection{Experiment Setup}
{\bf Baseline Methods}
We compare with two baselines:
\begin{itemize}
    \item \texttt{TCAFF} \cite{peterson2025tcaff}, which generates multiple hypotheses sequentially by removing correspondences obtained in previous runs and then re-solving the problem.
    \item \texttt{Stein ICP} \cite{maken2021stein}, which employs SVGD to perform mini-batch gradient descent on translational and rotational parts of the particles independently.
\end{itemize}
We also attempted to use Bayesian ICP \cite{maken2020estimating}, but it consistently diverged on our simulated object map examples.

\begin{figure}[t]
    \centering
    \includegraphics[width=0.48\textwidth,trim={0 .1cm 0 .2cm},clip]{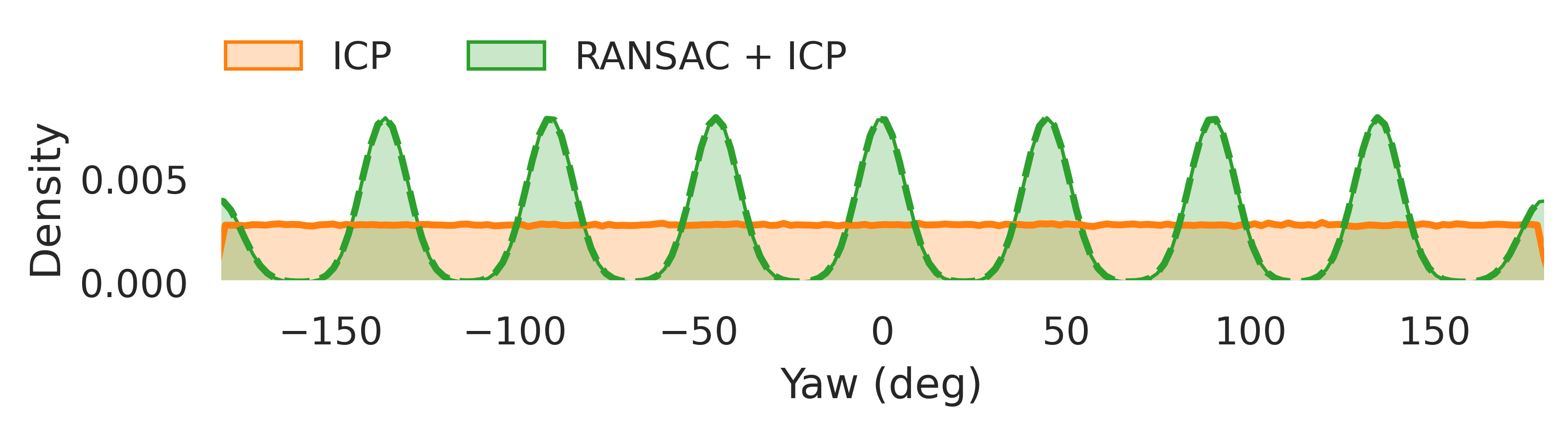}
    \caption{Visualization of Distribution on the Circle Example Generated by ICP vs. RANSAC proposals + ICP.}
    \label{fig:benchmark:circle_icp_ransac_icp}
\end{figure}

\label{sec:ref-dist}

{\bf {Ground Truth Distribution Generation}} 
In \cite{maken2021stein}, ground truth pose distributions are generated by running standard ICP from different initial guesses sampled from a small range ($\pm 1, \sim\pm 10 \deg$), which will not be able to capture the distributions usually encountered in object-level map association problems, where the transformations can spread all over $\mathrm{SE}(3)$. 
For example, consider the circle example shown in \Cref{fig:benchmark:exp_example} (top left). 
We run ICP from 1M initial poses by uniformly sampling rotations from $\mathrm{SO}(3)$ and translations in the bounding box defined by the extent of the map; the obtained yaw distribution is shown in  \Cref{fig:benchmark:circle_icp_ransac_icp} (yellow). 
We observe that the distribution is almost uniform. However, the peaks of the yaw distribution should only appear at multiples of $45 \deg$. 
One approach would be to enumerate all maximal cliques using Bron–Kerbosch (BK)~\cite{bron1973algorithm}; however, this is intractable for all but the smallest of problems.
Instead, we use the RANSAC-based alignment method described in \cite{aiger20084} to generate initial proposals by running it 1M times and keeping at most 5000 proposals. 
We then refine the accepted proposals with ICP \cite{besl1992icp}. The result is shown in \Cref{fig:benchmark:circle_icp_ransac_icp} (green). As we can see from the plot, this method cleanly captures the modes. The downside of this method is that it takes minutes to generate enough proposals to approximate the distribution. However, this is sufficient for the purpose of approximating the ground truth distribution offline.




{\bf Metric} 
Maken et al. \cite{maken2020estimating} used KL divergence on $x$, $y$, $z$, $\mathrm{roll}$, $\mathrm{pitch}$, and $\mathrm{yaw}$ separately, which may not represent the distribution faithfully because rotations live on $\mathrm{SO}(3)$ instead of $\R^3$. Followup work \cite{maken2021stein} instead reports a single number for KL divergence, but they do not detail their implementation. 
To the best of our knowledge, there is no standard method to numerically approximate KL divergence on $\mathrm{SE}(3)$. 
Instead, to better capture the discrepancy between distributions over $\mathrm{SE}(3)$, we use three different metrics: the Energy Distance (ED) \cite{rizzo2016energy}, Maximum Mean Discrepancy (MMD) \cite{gretton2012kernel}, and Wasserstein-1 distance (W1) \cite{cuturi2013sinkhorn}. For all three, we use Euclidean distance as the distance function on translation and chordal distance \cite{hartley2013rotation} as the distance function on rotation.

{\bf Implementation and Parameters} 
Since \texttt{Stein ICP} requires upper and lower bounds to sample initial poses from, for object map experiments, we set all the bounds on translations to be $0.5$m away from the bounding box enclosing the object maps. For the object point cloud examples, we sample translations from the unit cube, since the point clouds are rescaled to be within the unit cube. Rotations are sampled from $[-180 \deg, 180 \deg)$ to cover all possibilities.
We use 1000 particles and 1000 iterations for both \texttt{Stein ICP} and the proposed methods
to ensure the generated distributions are representative of the methods' capabilities. 
For {\texttt{TCAFF}}, since it discovers different solutions sequentially, we only generated 100 solutions to avoid long run times. 
The learning rates of \texttt{Stein ICP}, \texttt{Stein CLIPPER}, and {\texttt{Langevin CLIPPER}} are $0.01$, $0.001$, and $1.0$ respectively, which were obtained by tuning them on the two lines of points example. 
The AdaGrad parameters for the proposed methods are both set to $0.9$. The kernel bandwidth of \texttt{Stein CLIPPER} is set to $0.005$. The $\sigma$ and $\varepsilon$ for generating the affinity matrix $M$ are set to $(0.1$m$, 0.2$m$)$ for the object point cloud examples and $(0.4$m$, 0.6$m$)$ for everything else. Both proposed methods are implemented with PyTorch \cite{paszke2019pytorch}.

\begin{figure}[t]
    \centering
    \includegraphics[width=0.49\textwidth]{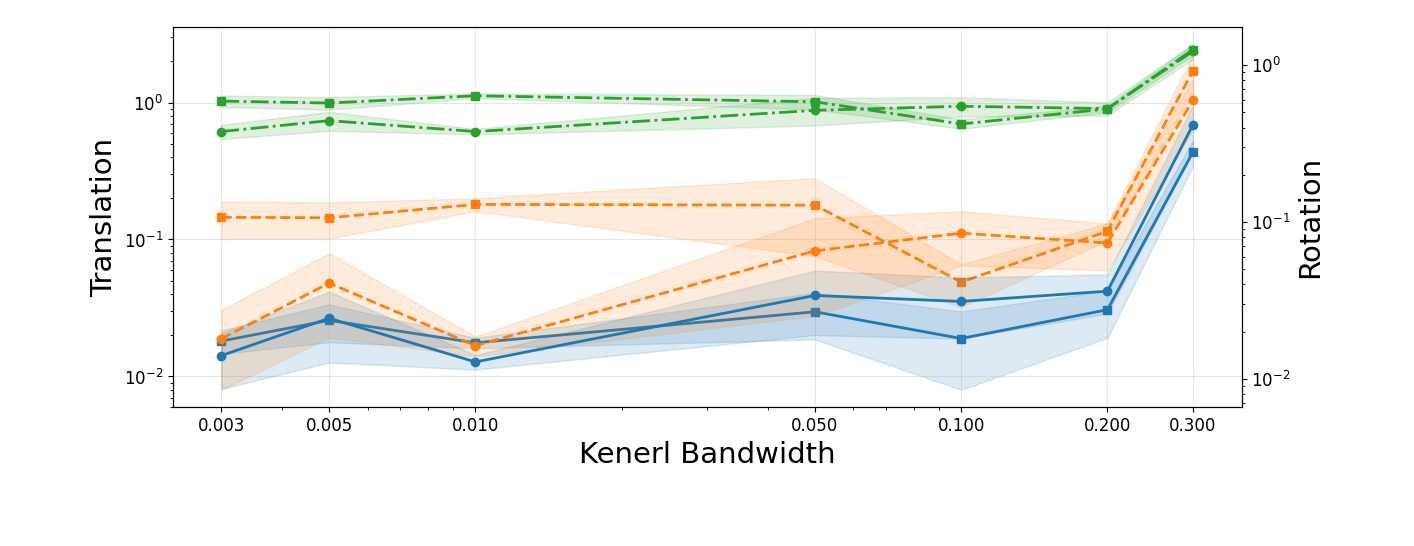}
    \caption{Ablations of \texttt{Stein CLIPPER} (right) over kernel bandwidth on the office example. We note that it fails to return solutions for bandwidth $< 0.003$ or $\ge 0.5$.}
    \label{fig:benchmark:ablation_stein_kernel_bandwidth}
\end{figure}

\begin{figure}[t]
    \centering
    \includegraphics[width=0.49\textwidth]{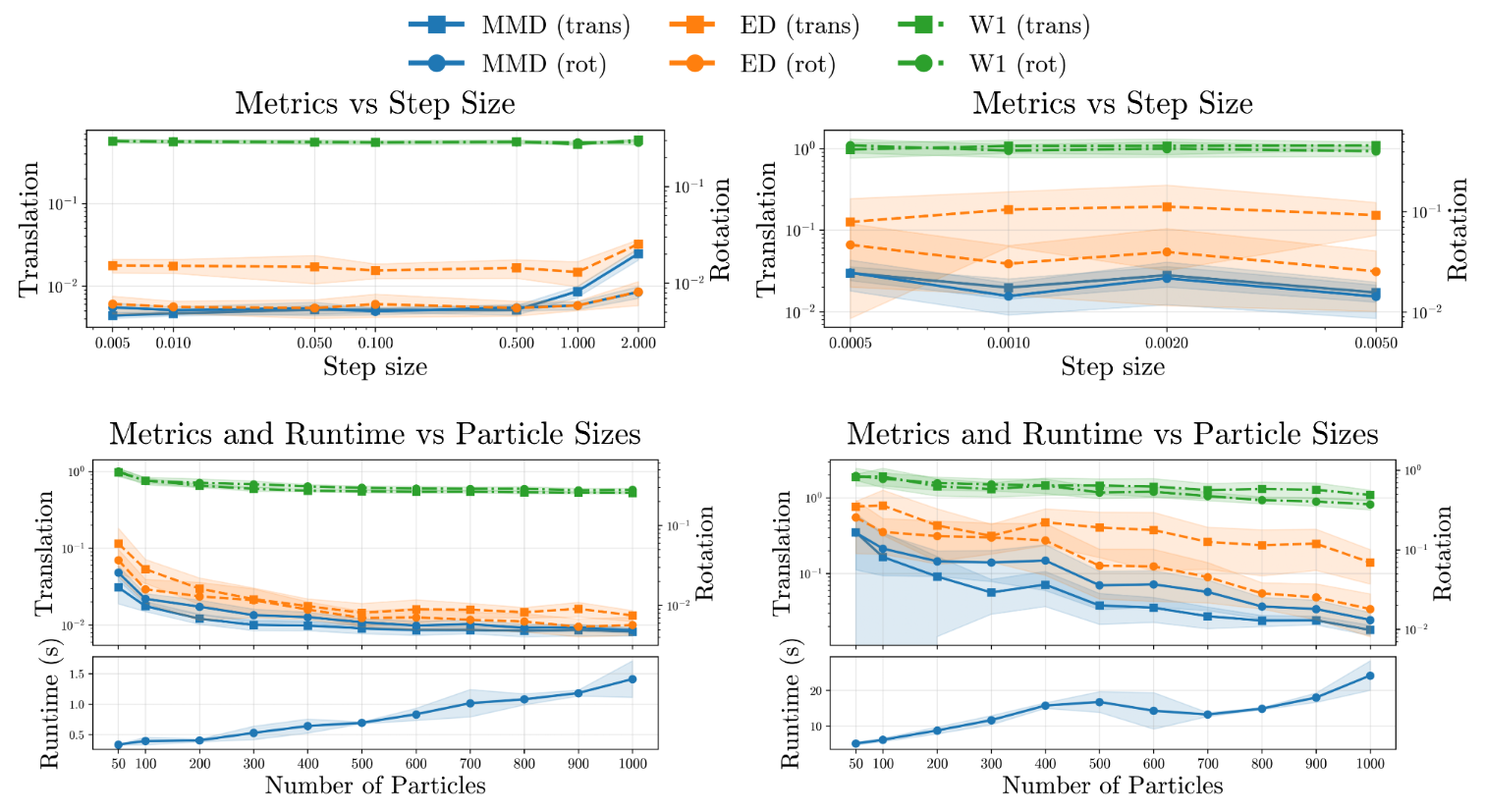}
    \caption{Ablations of \texttt{Langevin CLIPPER} (left) and \texttt{Stein CLIPPER} (right) over particle sizes and step sizes on the office example. We can see that \texttt{Langevin CLIPPER} is less sensitive to the parameters and can work with a larger range of step size.}
    \label{fig:benchmark:ablation_langevin}
\end{figure}

{\bf Ablation over Parameters}
An ablation study of the sensitivity of the proposed methods to the particle size and learning rate is performed using the office example, and the result is recorded in \Cref{fig:benchmark:ablation_langevin}. We find {\texttt{Langevin CLIPPER}} is less sensitive to the these parameters compared to \texttt{Stein CLIPPER}.
The ablation of \texttt{Stein CLIPPER} on the kernel bandwidth is recorded in \Cref{fig:benchmark:ablation_stein_kernel_bandwidth}. We note that in general the performance of \texttt{Stein CLIPPER} is sensitive to the choice of the kernel bandwidth. On one hand, a large kernel bandwidth may be insufficient to spread out particles. On the other hand, a small kernel bandwidth encourages particles to spread out but can cause aggressive updates that send particles outside of the feasible regions.

\subsection{Simulated Object Maps}
\label{sec:sim-obj-map-reg}
We report distribution similarity and runtime metrics in~\Cref{tab:sim-results}.
These values reflect the mean and standard deviation of running each method 10 times.

\begin{figure}[t]
    \centering
    \includegraphics[width=0.48\textwidth, trim=28 0 0 0, clip]{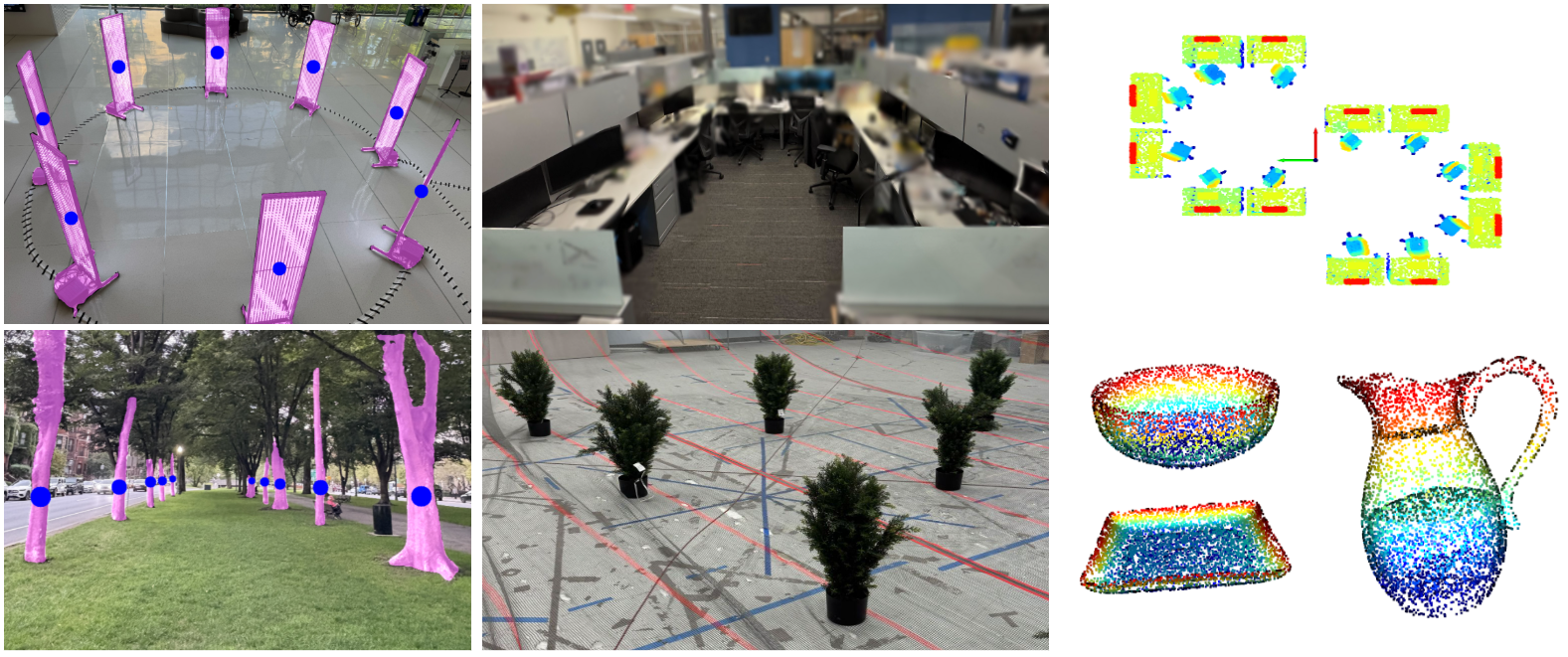}
    \caption{Visualizations of some examples used in experiments. For the examples on the left, we consider object maps formed by the highlighted objects. In the middle, the picture on the top showcases an office environment, which inspires us to create a simulated object map of the office (top right). The bottom picture of the middle column shows the example we use to study the effectiveness of the proposed approach in enumerating maximal cliques. The bottom right picture shows the object point clouds we use in \Cref{sec:sim_obj_pcd}.}
    \label{fig:benchmark:exp_example}
\end{figure}

\begin{figure}[t]
    \centering
    \includegraphics[width=0.48\textwidth,trim={0. .2cm 0. 0.4cm},clip]{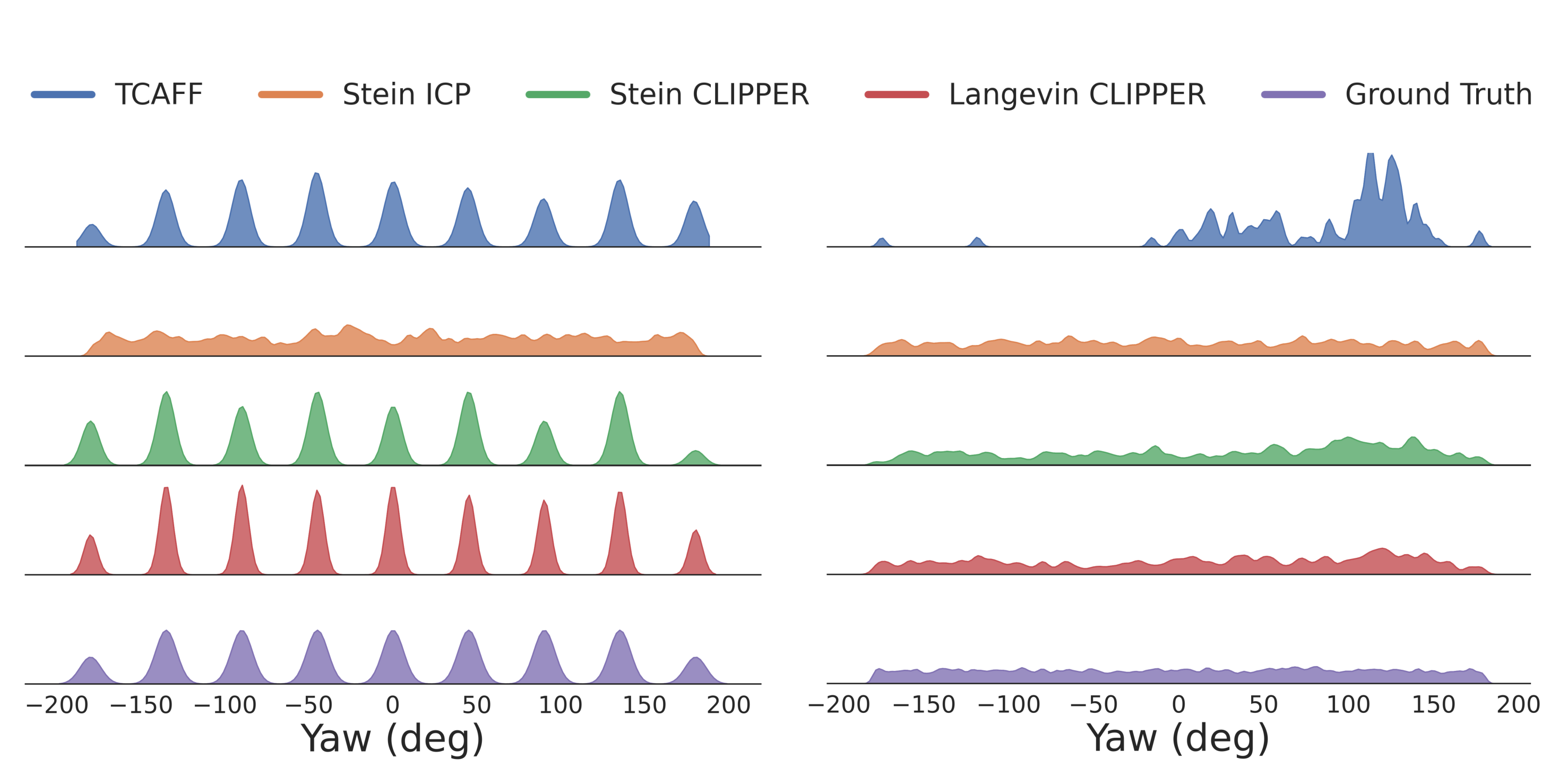}
    \caption{Comparisons of Yaw Distributions on Circle (left) and Bowl (right). 
    The proposed methods can handle both highly-peaked and uniform distributions.}
    \label{fig:benchmark:circle_bowl_comparison}
    \vspace{-0.6cm}
\end{figure}

{\bf Circle}
The first simulated object map is a circle of eight points lying on the $xy$ plane, motivated by \Cref{fig:benchmark:exp_example} (top left).
The yaw distributions obtained from all methods are shown in \Cref{fig:benchmark:circle_bowl_comparison} (left). \texttt{TCAFF} and the proposed methods successfully capture this discrete symmetry while \texttt{Stein ICP} struggles.
Quantitative results are recorded in \Cref{tab:sim-results} (first block). \texttt{TCAFF} and the proposed methods have low rotation errors, matching the qualitative results.

{\bf Two Lines of Points}
Motivated by \Cref{fig:benchmark:exp_example} (bottom left),
this configuration tests the ability of the proposed methods to capture a translational distribution with discrete modes.
\texttt{Langevin CLIPPER} and \texttt{TCAFF} are better at capturing the translation distribution exhibited in this example.  

{\bf Office}
The third simulated object map is an office, shown on the right of the top row in \Cref{fig:benchmark:exp_example}, where the object meshes are taken from ModelNet \cite{wu20153d}. We built two U-shaped pods, each consisting of 6 sets of workstations (including a chair, desk, keyboard, and monitor). This setup is motivated by the real-world office setup in \Cref{fig:benchmark:exp_example} (top middle). 
The quantitative results are recorded in the third block in \Cref{tab:sim-results}. As we can see, the proposed methods obtain best results overall, although \texttt{Stein CLIPPER} exhibits longer runtime. \texttt{TCAFF}'s runtime greatly increases in this example, which is expected due to the increase in the number of objects and the sequential computation of \texttt{TCAFF}.

\raggedbottom

\raggedbottom

\begin{table}[t]
    \scriptsize
    \centering
    \caption{Simulation Results}
    \setlength{\tabcolsep}{3pt}
    \renewcommand{\arraystretch}{1.1}
    \begin{tabular}{c l c c c c}
        \toprule
        & \textbf{Metric} 
        & \textbf{\texttt{TCAFF-100}} 
        & \textbf{\texttt{Stein ICP}} 
        & \shortstack{\textbf{\texttt{Stein}} \\ \textbf{\texttt{CLIPPER}}} 
        & \shortstack{\textbf{\texttt{Langevin}} \\ \textbf{\texttt{CLIPPER}}} \\
\midrule

        \multicolumn{6}{c}{\textbf{Object Map Registration}} \\
        \midrule

        \multirow{7}{*}{\rotatebox{90}{\textbf{Circle}}}
        & Runtime (s) & \best 0.04 ± 0.00 & 3.46 ± 0.22 & 1.77 ± 0.03 & \secbest 0.24 ± 0.01 \\
        & MMD (trans) & 2.29 ± 0.15 & 90.81 ± 1.74 & \best 0.00 ± 0.00 & \secbest 0.00 ± 0.00 \\
        & MMD (rot) & \secbest 0.19 ± 0.04 & 3.88 ± 0.22 & 1.19 ± 0.28 & \best 0.06 ± 0.02 \\
        & ED (trans) & 6.49 ± 0.18 & 84.07 ± 3.05 & \secbest 0.00 ± 0.00 & \best 0.00 ± 0.00 \\
        & ED (rot) & \secbest 0.59 ± 0.03 & 8.06 ± 0.24 & 2.12 ± 0.57 & \best 0.10 ± 0.04 \\
        & W1 (trans) & 13.81 ± 0.41 & 152.74 ± 3.92 & \secbest 0.00 ± 0.00 & \best 0.00 ± 0.00 \\
        & W1 (rot) & \secbest 12.23 ± 0.39 & 78.86 ± 1.38 & 23.07 ± 3.12 & \best 4.79 ± 0.78 \\
        \midrule

        \multirow{7}{*}{\rotatebox{90}{\textbf{Two Lines}}}
        & Runtime (s) & \secbest 0.97 ± 0.05 & 3.65 ± 0.21 & 3.10 ± 0.08 & \best 0.33 ± 0.02 \\
        & MMD (trans) & 1.94 ± 0.06 & \secbest 1.73 ± 0.47 & 12.74 ± 5.37 & \best 1.63 ± 0.11 \\
        & MMD (rot) & 7.58 ± 0.19 & 6.26 ± 0.15 & \secbest 4.68 ± 1.07 & \best 2.85 ± 0.14 \\
        & ED (trans) & \secbest 7.33 ± 0.44 & 21.19 ± 3.28 & 33.50 ± 14.32 & \best 4.66 ± 0.38 \\
        & ED (rot) & 5.88 ± 0.17 & 7.77 ± 0.25 & \secbest 4.24 ± 0.82 & \best 2.39 ± 0.10 \\
        & W1 (trans) & \secbest 78.58 ± 1.23 & 117.49 ± 5.50 & 127.85 ± 21.88 & \best 62.94 ± 1.28 \\
        & W1 (rot) & 52.46 ± 1.04 & 97.25 ± 0.86 & \secbest 41.70 ± 3.88 & \best 31.75 ± 0.86 \\
        \midrule

        \multirow{7}{*}{\rotatebox{90}{\textbf{Office}}}
        & Runtime (s) & 42.70 ± 0.88 & \secbest 3.70 ± 0.27 & 17.14 ± 0.30 & \best 1.19 ± 0.01 \\
        & MMD (trans) & 33.13 ± 1.76 & 2.52 ± 0.13 & \secbest 2.10 ± 0.37 & \best 0.77 ± 0.09 \\
        & MMD (rot) & 17.14 ± 1.24 & 3.16 ± 0.36 & \secbest 1.67 ± 0.66 & \best 0.57 ± 0.03 \\
        & ED (trans) & 61.71 ± 2.54 & 25.10 ± 0.83 & \secbest 12.36 ± 7.59 & \best 1.48 ± 0.38 \\
        & ED (rot) & 20.16 ± 2.29 & 4.52 ± 0.46 & \secbest 2.60 ± 1.44 & \best 0.63 ± 0.15 \\
        & W1 (trans) & 167.05 ± 2.55 & 120.30 ± 1.62 & \secbest 92.50 ± 15.90 & \best 53.63 ± 2.01 \\
        & W1 (rot) & 85.00 ± 3.78 & 52.21 ± 1.45 & \secbest 39.85 ± 6.54 & \best 28.99 ± 1.96 \\
        \midrule

        \multicolumn{6}{c}{\textbf{Point Cloud Registration}} \\
        \midrule

        \multirow{7}{*}{\rotatebox{90}{\textbf{Bowl}}}
        & Runtime (s) & 119.27 ± 4.00 & \secbest 4.10 ± 0.21 & 31.23 ± 0.81 & \best 0.83 ± 0.01 \\
        & MMD (trans) & 8.66 ± 0.11 & \secbest 5.25 ± 0.52 & 10.12 ± 3.57 & \best 3.49 ± 0.25 \\
        & MMD (rot) & 9.47 ± 0.31 & \secbest 0.88 ± 0.13 & 1.26 ± 0.28 & \best 0.83 ± 0.07 \\
        & ED (trans) & 5.56 ± 0.12 & \secbest 2.64 ± 0.34 & 5.58 ± 1.87 & \best 1.74 ± 0.15 \\
        & ED (rot) & 18.22 ± 0.93 & \secbest 4.74 ± 0.09 & 5.93 ± 1.52 & \best 1.38 ± 0.22 \\
        & W1 (trans) & 21.25 ± 0.07 & \secbest 18.80 ± 0.37 & 22.60 ± 2.35 & \best 17.14 ± 0.25 \\
        & W1 (rot) & 91.03 ± 0.91 & \secbest 55.03 ± 0.71 & 56.80 ± 3.78 & \best 43.05 ± 0.94 \\
        \midrule

        \multirow{7}{*}{\rotatebox{90}{\textbf{Plate}}}
        & Runtime (s) & 75.56 ± 1.74 & \secbest 4.15 ± 0.22 & 25.65 ± 1.62 & \best 0.69 ± 0.02 \\
        & MMD (trans) & \secbest 9.45 ± 0.13 & 9.64 ± 0.56 & 19.93 ± 11.65 & \best 4.10 ± 0.44 \\
        & MMD (rot) & 9.15 ± 0.28 & \secbest 0.85 ± 0.13 & 2.57 ± 0.59 & \best 0.81 ± 0.20 \\
        & ED (trans) & \secbest 4.53 ± 0.08 & 4.55 ± 0.37 & 9.12 ± 5.47 & \best 1.81 ± 0.18 \\
        & ED (rot) & 23.37 ± 1.17 & \best 0.47 ± 0.08 & 6.82 ± 3.17 & \secbest 1.61 ± 0.47 \\
        & W1 (trans) & \secbest 20.74 ± 0.10 & 21.07 ± 0.27 & 28.89 ± 5.93 & \best 16.96 ± 0.32 \\
        & W1 (rot) & 79.83 ± 1.28 & \best 27.33 ± 0.89 & 51.20 ± 7.99 & \secbest 32.98 ± 2.64 \\
        \midrule

        \multirow{7}{*}{\rotatebox{90}{\textbf{Pitcher}}}
        & Runtime (s) & 212.77 ± 3.51 & \secbest 4.35 ± 0.20 & 31.76 ± 0.15 & \best 1.04 ± 0.01 \\
        & MMD (trans) & 6.95 ± 0.15 & \secbest 6.22 ± 0.75 & 55.54 ± 8.50 & \best 2.90 ± 0.25 \\
        & MMD (rot) & 6.50 ± 0.34 & \secbest 1.20 ± 0.40 & 3.41 ± 0.38 & \best 0.51 ± 0.05 \\
        & ED (trans) & 3.31 ± 0.11 & \secbest 2.85 ± 0.39 & 40.61 ± 5.81 & \best 1.22 ± 0.11 \\
        & ED (rot) & 35.40 ± 0.85 & \best 2.84 ± 1.04 & 23.60 ± 3.69 & \secbest 3.15 ± 0.54 \\
        & W1 (trans) & 20.72 ± 0.11 & \secbest 19.59 ± 0.57 & 48.18 ± 2.66 & \best 16.98 ± 0.22 \\
        & W1 (rot) & 116.70 ± 0.93 & \secbest 54.09 ± 5.11 & 85.16 ± 4.24 & \best 47.89 ± 1.75 \\


        \bottomrule
    \end{tabular}
    \footnotesize{All values except runtimes are multiplied by $10^2$ for readability.}
    \label{tab:sim-results}
\end{table}
\raggedbottom

\subsection{Object Point Clouds}\label{sec:sim_obj_pcd}
To investigate the performance of the proposed methods on registering object point clouds, we choose three object scans from the Google Scanned Objects dataset \cite{downs2022google}, shown in \Cref{fig:benchmark:exp_example} (bottom right). This set of experiments is motivated by the experiment setup in \texttt{Stein ICP} \cite{maken2021stein}.
The point clouds are rescaled to a unit cube and then downsampled with voxel size $0.05$m. Noise is injected from $\mathcal{N}(0, 0.025)$. We use FPFH \cite{rusu2009fast} features to generate putative associations for \texttt{TCAFF} and the proposed methods. A visualization of the yaw distributions for the bowl is provided in \Cref{fig:benchmark:circle_bowl_comparison} (right). We observe that \texttt{Stein ICP} and \texttt{Langevin CLIPPER} are better at capturing the distribution 
which is more uniform than those in the object map examples. 
All the quantitative results are recorded in the lower half of \Cref{tab:sim-results}. We observe that \texttt{Langevin CLIPPER} performs best overall, which shows the ability of \texttt{Langevin CLIPPER} to capture both highly-peaked distributions and uniform distributions. 
\texttt{Stein ICP} performs better on this set of experiments than on object map registrations, as it was developed and evaluated on similar scenarios.
\texttt{Stein CLIPPER} performs worse on this set of experiments, which we believe is due to the kernel bandwidth. In practice, one could use different bandwidths for different experiments, but we keep the bandwidth the same to test generalizability.

\section{Real World Experiments}
\label{sec:hardware_experiment}
To evaluate the capabilities of the proposed methods in scenarios with real sensor noise, we perform real-world object map registration experiments. Note that we use the {\it same parameters} for proposed methods as we used for simulated object maps throughout this section.

\subsection{3RScan Dataset}
We evaluate the proposed methods using the 3RScan dataset
\cite{Wald2019RIO}, which contains scans of indoor scenes captured with RGB-D cameras. 
This dataset tests whether the proposed methods can be applied to general scenes, irrespective of the multimodal nature of the underlying data association distribution, as the 3RScan scenes may not necessarily elicit multimodal association solutions.
We randomly select 100 scenes from 3RScan, where 2 scans are randomly selected from each scene for registration. We use the same ground truth generation as in the simulation experiments.
The results averaged over all 100 scenes are recorded in \Cref{tab:3rscan_result}.
An interesting observation is that \texttt{TCAFF} performs better on this dataset than it does in the simulation study, despite having a much longer runtime than the other methods. We hypothesize this is due to the lack of ambiguous data in 3RScan, in which case \texttt{TCAFF} becomes similar to a standard data association method. \texttt{Langevin CLIPPER} still performs the best overall, demonstrating its effectiveness in general data association problems, regardless of whether the solution distribution is unimodal or multimodal.


\begin{table}[t]
    \scriptsize
    \centering
    \caption{3RScan Dataset Results}
    \setlength{\tabcolsep}{3pt}
    \renewcommand{\arraystretch}{1.1}
    \begin{tabular}{c l c c c c}
        \toprule
        & \textbf{Metric} 
        & \textbf{\texttt{TCAFF-100}} 
        & \textbf{\texttt{Stein ICP}} 
        & \shortstack{\textbf{\texttt{Stein}} \\ \textbf{\texttt{CLIPPER}}} 
        & \shortstack{\textbf{\texttt{Langevin}} \\ \textbf{\texttt{CLIPPER}}} \\
\midrule
        & Runtime (s) & 13.42 ± 26.15 & 3.52 ± 0.15 & \secbest 2.61 ± 1.45 & \best 0.37 ± 0.13 \\
        & MMD (trans) & 2.45 ± 1.58 & \secbest 1.64 ± 1.59 & 3.09 ± 3.87 & \best 1.16 ± 0.98 \\
        & MMD (rot) & 2.37 ± 1.66 & \secbest 2.02 ± 2.46 & 2.66 ± 3.98 & \best 1.20 ± 1.32 \\
        & ED (trans) & \secbest 6.40 ± 6.63 & 7.63 ± 10.62 & 38.50 ± 54.70 & \best 4.12 ± 5.83 \\
        & ED (rot) & \secbest 4.11 ± 3.66 & 4.33 ± 5.45 & 9.80 ± 15.21 & \best 2.57 ± 2.99 \\
        & W1 (trans) & \secbest 63.52 ± 22.55 & 68.24 ± 30.77 & 104.40 ± 63.26 & \best 51.08 ± 23.24 \\
        & W1 (rot) & \secbest 58.86 ± 9.22 & 59.78 ± 15.08 & 65.28 ± 20.65 & \best 46.01 ± 9.46 \\
        \bottomrule
    \end{tabular}
    \footnotesize{All values except runtimes are multiplied by $10^2$ for readability.}
    \label{tab:3rscan_result}
\end{table}
\raggedbottom


\subsection{Real World Object Map Registration}

To demonstrate the importance of understanding uncertainty in ambiguous association problems,
we apply our method to several real-world scenarios in which ROMAN~\cite{peterson2024roman}, a state-of-the-art object map association method, commits to the wrong mode of the solution distribution.
%
Initial object maps are built online using ROMAN, which processes RGBD images from a RealSense D455 stereo camera with Kimera-VIO~\cite{rosinol2020kimera} for odometry.
We directly use the affinity matrix $M$ provided by ROMAN, which incorporates gravity direction and object semantic similarity priors, but fails to uniquely disambiguate between repetitive objects.
Data is recorded from the three ambiguous environments shown in~\Cref{fig:demo}: a staircase, a line of trees, and a circular library room.
We task each algorithm with registering two object maps: one complete reference map and a smaller second map (e.g., a single floor of the staircase).
For each experiment, we investigate an ambiguous axis of the estimated pose.
We visualize the estimated pose distribution found by the proposed methods, the pose found by ROMAN, and a manually annotated ground truth pose for qualitative comparison.
We find that \texttt{Langevin CLIPPER} successfully represents the multimodal distribution while \texttt{Stein CLIPPER} often collapses to a few modes, likely due to its sensitivity to the kernel bandwidth.

\begin{figure}[t]
    \centering
    \includegraphics[width=0.48\textwidth]{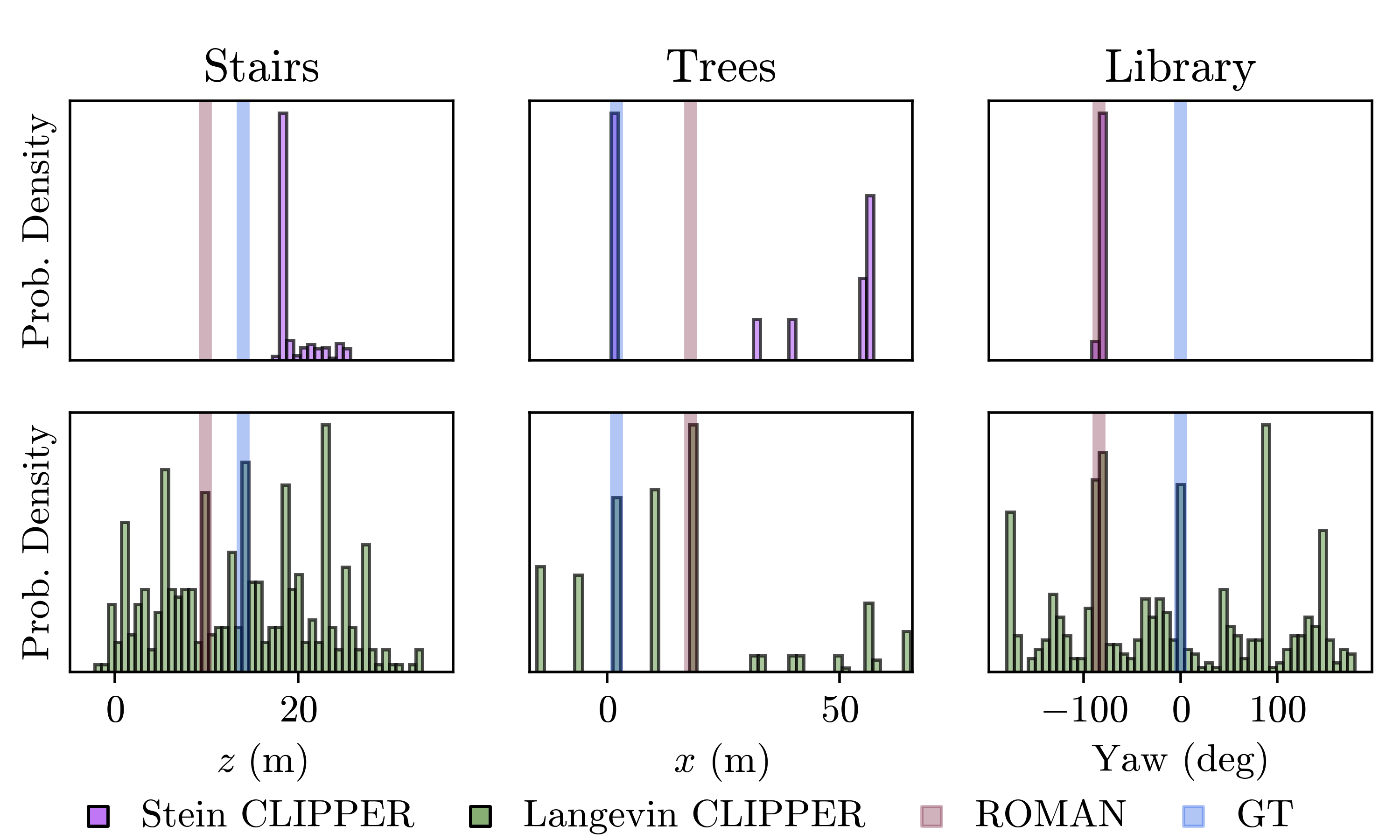}
    \caption{Pose distributions from \texttt{Stein CLIPPER} and \texttt{Langevin CLIPPER} for the staircase, trees, and library shown in~\Cref{fig:demo}. For each case, the axis containing repetitive geometry is shown. \texttt{Langevin CLIPPER} successfully finds modes at different floors of the staircase, at different trees in the line, and around the four quadrants of the library.
    The single estimate found by ROMAN and the ground truth (GT) pose are also shown. 
    }
    \label{fig:hw}
\end{figure}

{\bf Stairs}
Significant elevation ($z$-axis) ambiguity is present in this example, which is shown in \Cref{fig:demo} (left) with distributions generated by proposed methods visualized \Cref{fig:hw} (left).
\texttt{Stein CLIPPER} fails to capture the full distribution and ROMAN commits to an incorrect mode. 
Meanwhile, \texttt{Langevin CLIPPER} recovers a peak at elevations corresponding to each floor of the stairway, capturing the uncertainty of the solution.

{\bf Trees}
In this example, shown in \Cref{fig:demo} (middle), a single-file line of tall trees is mapped, inducing ambiguity in the pose estimate's $x$-axis.
\texttt{Stein CLIPPER} and \texttt{Langevin CLIPPER} both capture many potential modes at regular intervals, although \texttt{Langevin CLIPPER} finds more modes.

{\bf Library}
Our final example, shown in \Cref{fig:demo} (right), exhibits yaw ambiguity. 
The library room is circular, with four doors and columns between each door. 
\texttt{Stein CLIPPER} and ROMAN both commit to the wrong mode, while \texttt{Langevin CLIPPER} finds the four modes at -90, 0, 90, and 180 deg.

\subsection{Enumerating Maximal Cliques}\label{subsec:bk_result}
\begin{figure}[t]
    \centering
    \begin{subfigure}{0.48\textwidth}
        \centering
        \includegraphics[width=\textwidth,trim={0 .3cm 0 .3cm},clip]{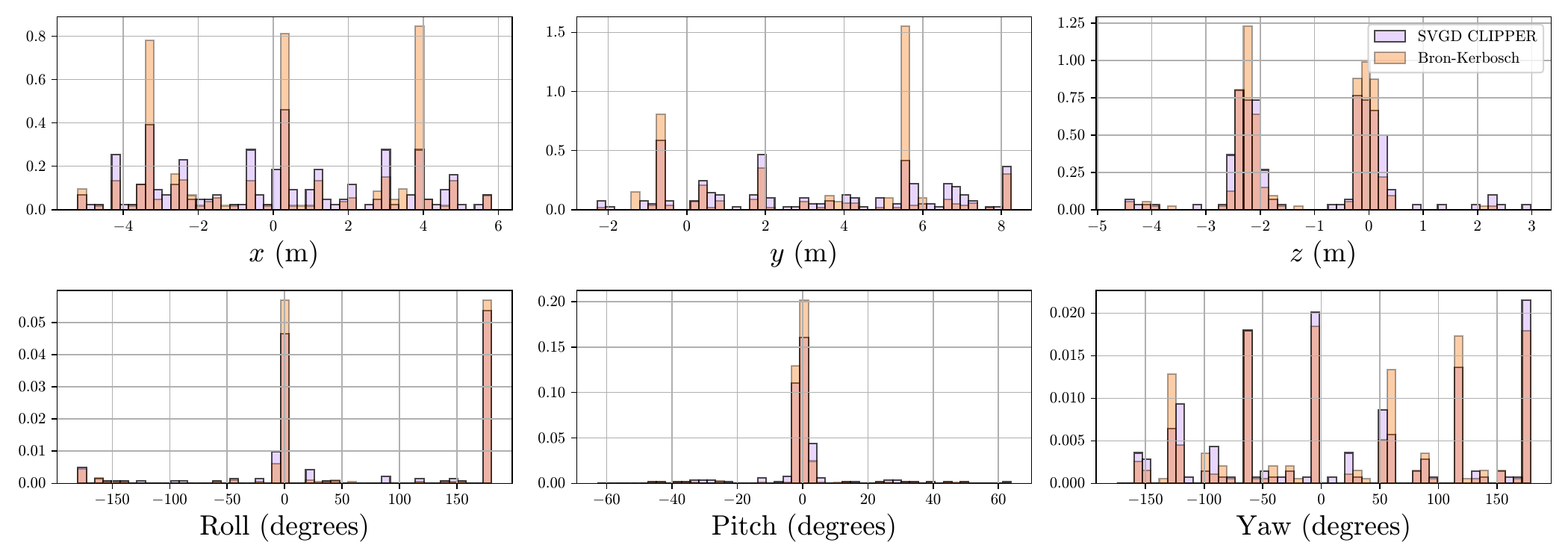}
        \label{fig:bk_ld_a}
    \end{subfigure}
    
    \vspace{-0.3cm} 
    
    \begin{subfigure}{0.48\textwidth}
        \centering
        \includegraphics[width=\textwidth,trim={0 .3cm 0 .3cm},clip]{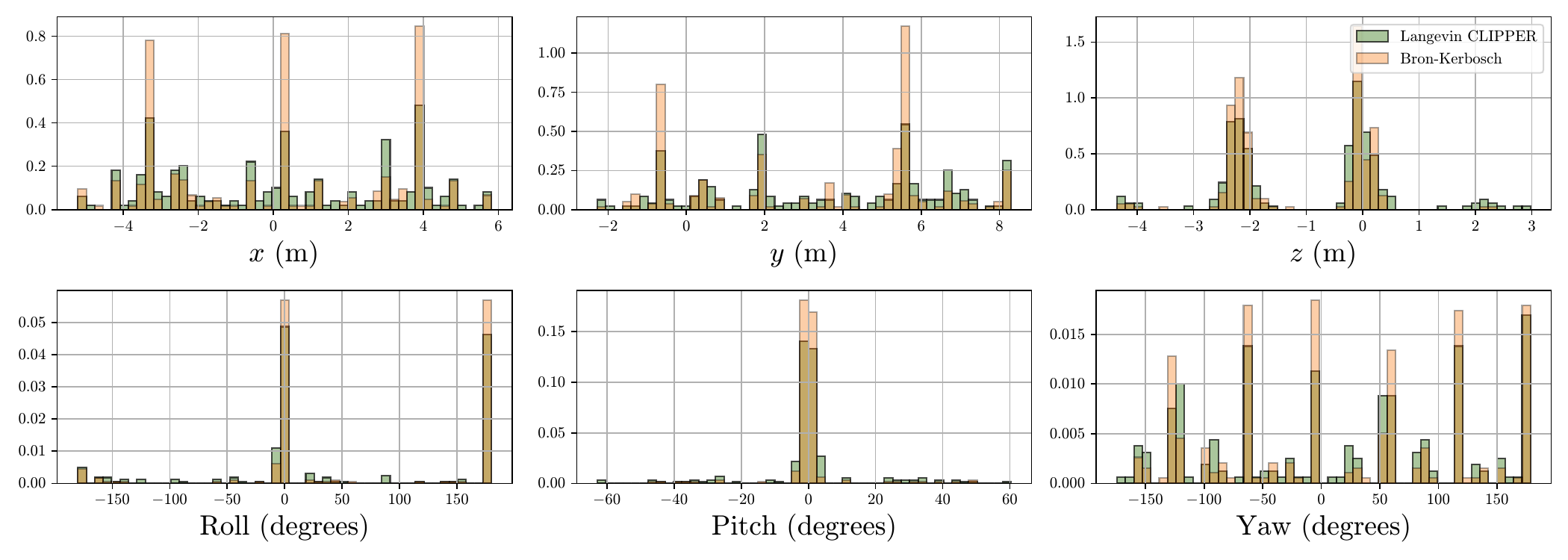}
        \label{fig:bk_ld_b}
    \end{subfigure}
    \vspace{-0.8cm}
    
    \caption{Visualization of the theoretical distribution (Bron–Kerbosch) versus the actual distributions generated by the proposed methods on the toy example shown in \Cref{fig:benchmark:exp_example} (bottom row middle). We observe that both methods capture the modes, despite being less concentrated. Note that the two object maps were taken from different views, thus the non-zero modes in translation.}
    \label{fig:bk_ld}
\end{figure}

We set up a small example to investigate how well the proposed methods match the theoretical result discussed in \Cref{rem:bk}, where enumerating all maximal cliques is possible using the BK \cite{bron1973algorithm} algorithm. The setup of the experiment is shown in \Cref{fig:benchmark:exp_example} (bottom middle), where two object maps are generated 
from two different views of the triangular arrangement of bushes.
We run BK to generate all the maximal cliques of the consistency graph, by rounding edge weights above $0.5$ to 1 and others to $0$. The result is shown in \Cref{fig:bk_ld}. We can see that both proposed methods matches BK well, although the resulting distribution is less concentrated around the modes (i.e., smaller density), which is expected since the theoretical analysis is asymptotic.



\section{Conclusion}
In this work, we proposed and investigated two methods for capturing multimodal distributions of solutions to global data association problems, leveraging approximate Bayesian inference techniques.  
We showed that Langevin CLIPPER is able to capture both highly-peaked and more uniform distributions, in simulation and with noisy real-world data. One potential downstream application is uncertainty estimation in loop closure and pose estimation problems.

\bibliographystyle{IEEEtran}
\bibliography{paper/references}

\end{document}